# Context Reasoning in Underwater Robots Using MEBN

Xin Li, José-Fernán Martínez, Gregorio Rubio and David Gómez
Centro de Investigación en Tecnologías Software y Sistemas Multimedia para la Sostenibilidad (CITSEM)
Universidad Politécnica de Madrid
Ctra. de Valencia, Km. 7 28031 – Madrid (Spain)
E-mail: xin.li@upm.es, jf.martinez@upm.es, gregorio.rubio@upm.es, david.gomezs@upm.es

*Abstract*—This paper presents ongoing research in the SWARMs project towards facilitating context awareness in underwater robots. In particular, the focus of this paper is put on the context reasoning part. The underwater environment introduces uncertainties in context data which lead to difficulties in the context reasoning phase. As probability is the best well-known formalism for computational scientific reasoning under uncertainties, the emerging and effective probabilistic reasoning method, namely, Multi-Entity Bayesian Network (MEBN), is explored for its feasibility to reason under uncertainties in the SWARMs project. A simple use case for oil spill monitoring is used to verify the usefulness of MEBN. The results show that the MEBN is a promising approach to reason about context in the presence of uncertainties in the underwater robot field.

## I. Introduction

The work presented in this paper is part of the SWARMs (Smart and Networking Underwater Robots in Cooperation Meshes) project (http://www.swarms.eu/), a European project in the area of underwater robotics. The aim of the SWARMs project is to expand the use of underwater robots (e.g., AUVs, ROVs) in maritime and offshore operations and enable them to work in a cooperative, coordinative, and context-aware manner. The SWARMs approach is underpinned by designing a semantic and distributed middleware. The middleware layer is basically designed to facilitate communication between heterogeneous vehicles and Mission Management Tool (MMT) which is located in Command & Control Station (C & CS) and provide a set of common services. The middleware features a context-aware framework which is dedicated to delivering context awareness in underwater robots.

Context awareness is a significant feature that could enable underwater robots to understand the complete picture of the underwater environment and adapt their behaviors accordingly. Context awareness implies an effective exploitation of context. To maximize the use of context, the context-aware framework is envisaged to provide well-defined context treatments, such as abstracting context heterogeneity, enabling sharing and reuse of context & capabilities between AUVs/ROVs, delivering useful information to robots & operators for decision making etc.

The services provided by the context-aware framework generally comply with the lifecycle of context awareness. According to Perera et al. [1], the lifecycle of context awareness mainly consists of four phases, including context acquisition, context modeling, context reasoning, and context dissemination. Especially, context reasoning is a significant process because underwater robots or operators are more willing to use high-level context for decision-making. For instance, a piece of high-level context information, *vehicle A might collide with vehicle B soon*, deduced from available context information *vehicle A is out of trajectory and heading in the direction of vehicle B*, could enable operators with a better understanding of the situation. In the SWARMs project, a variety of context, referring to any information that could be used to characterize underwater robots and their operational environment, could be obtained, such as battery level of vehicles, capabilities of vehicles, currents, turbidity, salinity, and wind direction etc. Considering the harsh nature of underwater environments, context is prone to be uncertain due to unreliable communications, data loss, imperfect instruments, or partial views, etc [2]. The uncertainty, as an inherent characteristic of context, introduces more challenges in context reasoning in the underwater robot field.

Ontological reasoning and rule-based reasoning are widely used context reasoning methods. However, they lack the capability of reasoning under uncertainties as they assume a deterministic world. Probability is the best well-known formalism for computational scientific reasoning under uncertainties. As a widely used probabilistic reasoning method, Bayesian networks (BN) could represent and reason about problems involving many related hypotheses. However, in a standard BN, all the hypotheses and dependencies must be fixed and only the evidence could be varied from problem to problem. However, many real-world problems could involve a varying number of entities. Thus, standard BN cannot be able to handle uncertainties in such real problems. Multi-Entity Bayesian Network (MEBN) [3] extends BN to achieve first-order expressive power and it provides a compact way to represent repeated structures in a BN. MEBN has shown its usefulness in reasoning under uncertainties in many applications, such as procurement fraud detection [4], maritime awareness [5], knowledge-driven analysis for cultural heritage [6]. In the SWARMs project, there may exist a lot of complex reasoning problems which include a varying number of entities. For example, the investigated robot may have more than one robot around. The collision risk is affected by all possible nearby robots, thus, resulting in a problem which involves a varying number of entities. The MEBN shows the potential in reasoning about such complex problems in the SWARMs project. Therefore, it is worth exploring the usefulness of the MEBN in the SWARMs project. This paper presents the conceptual proposal of context-aware framework defined in the SWARMs project and explores the applicability of the MEBN in context reasoning using a case study. The usefulness of MEBN is verified by reasoning the severity level of a spilled region in an oil spill monitoring use case.

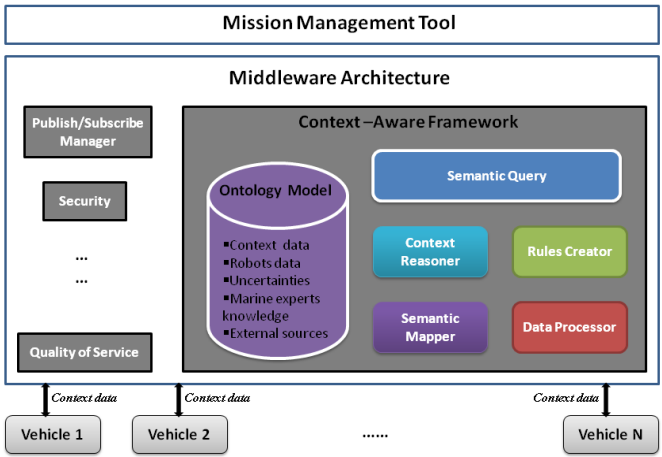

Fig. 1. The proposed context-aware framework.

## II. THE PROPOSED CONTEXT-AWARE FRAMEWORK

The context-aware framework, as part of the SWARMs middleware, is shown in fig. 1. It is conceived to be modular and distributed which could imply a better potential over a centralized one to be employed in robots coordination and cooperation.

The context-aware framework consists of six logic components, namely, *Data Processor*, *Semantic Mapper*, *Ontology Model*, *Rules Creator*, *Context Reasoner*, and *Semantic Query*. Functional capabilities of each component are described in the following sections.

*Data Processor*. The extreme underwater conditions impose more challenges in the data acquisition phase and context data obtained from the environment could be uncertain. A preliminary data treatment is provided by the data processor component. This component can pre-process data received from sensors or other sensing instruments. Statistics on historical data, such as probability distribution, can be learned using machine learning algorithms in this component. Operations, such as validating values, checking inconsistencies, calculating uncertainty degrees, removing outliers or filling in missing values etc., can be executed in this component.

*Ontology Model*. This component, acting as a semantic repository, keeps the SWARMs ontology and stores all data obtained as instances in the ontology model. Ontology is adopted to serve as a common information model to represent information and enable sharing, reuse, and integration of data between vehicles. The information model is structured in a hierarchical manner. The networked information model consists of three levels of ontologies: core ontology (acting as an upper-level ontology to glue all domain specific ontologies), domain specific ontology (providing information models for different domains, e.g., mission planning, robotic vehicles, environment recognition & sensing, and communications & networking), and application-specific ontology (describing information with a focus on particular applications). Therefore, data, including, but not limited to contextual measurements, vehicle-related data, data from external sources (oceanic weather forecast etc.), and marine experts knowledge, can be abstracted and formalized in an ontological format. All data can be displayed with a homogeneous view associated with the semantic content. The ontology model could annotate probability information of uncertain context by using ontology constructs defined in the PR-OWL [3] ontology. With the PR-OWL classes and properties, ontology engineers could fully specify an MEBN model using the OWL ontology language. Thus, the SWARMs ontology could not only annotate probability information about uncertain context but also provide support for the MEBN reasoning.

*Semantic Mapper*. This component plays a vital role in facilitating the transparent sharing of information. As data might be formatted in different manners pertaining to different data sources, this component aims to parse them and formalize them in an ontology compliant format. Translations from different standards, such as XML, JSON, or binary files to ontological formations (RDF, or OWL) can be enabled in this component. For differently formatted data which do not comply with the SWARMs ontology, corresponding mapping files should be predefined in order to parse and map them into the common information model.

*Rules Creator*. Operators or marine experts are able to define rules according to their knowledge and changes in the environment through this component. A set of user-defined rules can be translated into the SWRL (Semantic Web Rule Language) format and inserted into the ontology model. Rules can be diverse, including restrictions or definitions for entities, regulations for evaluating data values, or specifications for relationships between entities. The rule set is very important to be considered by the context reasoner to make inferences.

*Context Reasoner*. The essential capability of this component is to deduce new knowledge based on available context stored in the ontology model. Basically, it will consult information stored in the ontology model and also take experiences and knowledge from marine experts into account. A hybrid reasoning mechanism, including ontological, rule-based and MEBN reasoning, is intended to be adopted by this context reasoner. Specifically, ontological reasoning enables several kinds of operations, including concept satisfiability, consistency check, class subsumption and logic inference. The rule-based reasoning could augment ontological reasoning in terms of logicality and human-readability. The MEBN is dedicated to reasoning under uncertainties and it is the focus of this paper which will be elaborated in section III and IV.

*Semantic Query*. This component deals with any semantic query made by operators. It receives queries, calls corresponding reasoning services, and finally outputs answers. This is one of context dissemination strategies defined in the SWARMs project. Apart from this, the SWARMs project also follows a publishing/subscription paradigm to distribute context information.

The context-aware framework provides architectural support for different context treatments. In particular, the context reasoner is of significance as it is expected to provide necessary services to meet all the reasoning requirements in the SWARMs project. Especially, it should handle the intricacies adherent to the reasoning over uncertainties. The feasibility of integrating MEBN in the context reasoner is explored in the following sections.

## III. PRINCIPLES OF MEBN

MEBN unifies Bayesian probability and statistics with classical first-order logic. By incorporating Bayesian probability with first-order logic, the MEBN is able to provide a consistent treatment of uncertainty, such as representing uncertainty about the type of an entity, refining type-specific probability distributions through Bayesian Learning, and reasoning under uncertainties. Beyond the capability of traditional BN in reasoning about the fixed number of attributes, MEBN could deal with a varying number of entities with their number, type, and relationships undetermined. MEBN provides a means of defining probability distributions over an unbounded and varying number of interrelated hypotheses with the aid of syntax, a set of model construction and inference processes, and semantics.

MEBN interprets the world as a set of entities that have attributes and have causal relationships with other entities. Knowledge about the attributes of the entities and their relationships to each other is represented as an MEBN model. MEBN logic consists of a collection of MEBN fragments (MFrag) organized into an MEBN Theory (MTheory). An MTheory could represent a particular domain of discourse. Each MFrag, as a modular component, represents knowledge about specific subjects within the domain of discourse and it models probability information about a group of related random variables. Similar to BN, each MFrag is a Directed Acyclic Graph (DAG) with parameterized nodes that represent attributes of entity and edges that represent dependencies amongst them. Three types of nodes, including *resident nodes*, *context nodes*, *input nodes*, are defined in the MTheory.

- *Resident nodes.* In an MFrag graph, resident nodes represent variables that have local probability distributions dependent on the values of their parents. Exactly one home MFrag is assigned to contain the complete expression of a resident node.

- *Context nodes.* Context nodes are Boolean nodes, including value True, False, and Absurd. The MFrag must satisfy the conditions expressed by context nodes in order to be valid.

- *Input nodes.* Input nodes have their distributions defined in other MFrags and they are important inputs for the definition of resident nodes.

The MTheory could be instantiated with specific information about the individual entity instances to reason about specific situations. The instantiated MEBN, named Situation Specific Bayesian Network (SSBN), could make the inference under uncertainties based on standard BN reasoning. The open source UnBBayes [7] tool is available to provide both GUI and Java API for building MEBN and executing learning and reasoning.

## IV. USING MEBN IN AN OIL SPILL MONITORING PROBLEM

In this section, a case study on oil spill monitoring is presented. Through this case study, the comparison between the BN and the MEBN reasoning is provided and also the applicability of the MEBN reasoning is verified.

### A. Description of the Oil Spill Monitoring Scenario

Oil spill detection is considered as one of use cases in the SWARMs project. Proactive detection of oil spills is an important means to minimize the damage of spills to the marine environment. A set of SWARMs vehicles could collaboratively detect the occurrence of oil spills based on contextual data they could obtain. After detecting oil spills in a specific marine region, it is significant to predict the severity level of the spilled region so that operators could have a better understanding of the situation and make mission planning/re-planning (e.g., clean-up, containment) for underwater robots. The estimation of the severity level of the specific area is subject to several available contexts, such as the thickness of spills, the estimated size of spills, weather conditions, and currents. In addition, the severity level of the specific area could be influenced by different spills concurrently occurring in the same area.

### B. BN for Representing the Scenario

Fig. 2(a) is the BN that can be used to represent the severity level and its influencing variables. It expresses that currents, weather, thickness, and estimated size are the main criteria to determine the severity level of a spilled region. Fig. 2(a) shows the major shortage of the BN is that it cannot serve if there is any change to the scenario. For instance, if two oil spills (spill_1 and spill_2) are detected in a specific sea area (region_1). Spill_1 is observed as being thick and large and spill_2 is detected as being thin and small. In addition, the weather condition above the spilled marine region is known to be inclement. Currents in the region_1 fluctuate very strongly. The BN, shown in fig. 2(a), cannot accommodate to this situation which contains two spill instances. It is necessary to define a new BN, illustrated in fig. 2(b), to provide the corresponding representation and reasoning. When changes occur in the scenario (e.g., three, or more spills exist in the same region), BN models need to be revised. The BN, shown in fig. 2(b), repeatedly models elements from the BN shown in fig. 2(a) with the same semantics. From a domain modeling perspective, a more generic and flexible approach is needed to allow instantiation of nodes in order to adapt to different situation.

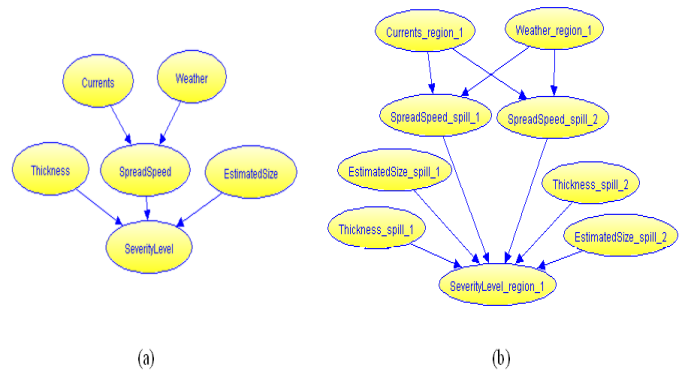

Fig. 2. Classical BNs to reason about the scenario.

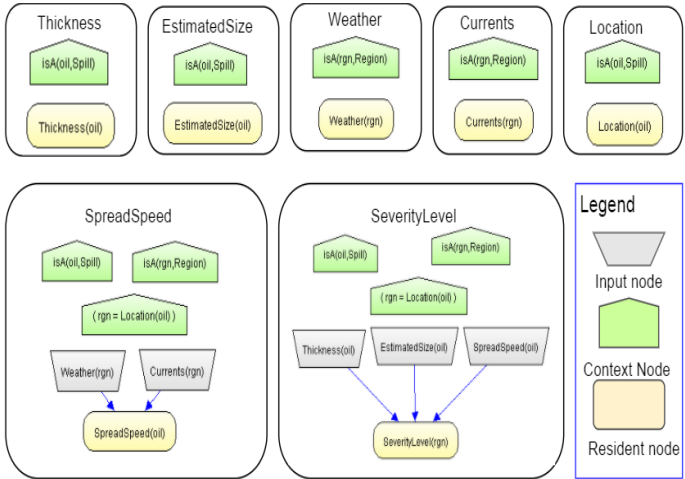

Fig. 3. The proposed MEBN model.

C. *MEBN for Representing the Scenario*

An MEBN model, shown in fig. 3, is built using the UnBBayes tool to represent the described reasoning problem in a modular way.

The MTheory consists of seven MFrags, including *Thickness*, *EstimatedSize*, *Weather*, *Currents*, *Location*, *SpreadSpeed*, and *SeverityLevel*. Each MFrag represents knowledge for a specific entity and its local probability distribution. For instance, the *SpreadSpeed* MFrag models that the spread speed of detected spills is affected by two inputs, namely, weather and currents. These two input nodes, weather and currents, are specified in the *Weather* and *Currents* MFrags, respectively. The local probability distribution of *SpreadSpeed* is shown in fig. 4. For the demonstration purpose, all the probability information in this model is directly provided by marine experts based on their experiences.

With the seven MFrags, the MTheory could collectively model the unique joint probability distribution for the *SeverityLevel* entity. This MTheory could be instantiated to specific scenarios and make inferences based on observed evidence correspondingly. For instance, given the same findings (shown in fig. 5) which are described in section B, the MEBN could be specialized to the described scenario, namely, SSBN, and it is shown in fig. 6.

With the evidence listed in fig. 5, the SSBN can answer specific questions of interest. For instance, based on standard BN reasoning, the SSBN could infer that the region_1 with the existence of two spills could be estimated as very serious with a probability of 88%. This high-level context could be sent to the MMT for further exploitation. For instance, it could be used by operators as a parameter to conceive a plan for underwater robots so that they can take remedial measures accordingly. In this situation, only two spill instances are exemplified as influence factors for the severity level. It is worth noting that the built MEBN could be flexible to accommodate to reasoning which involves more spill instances. Therefore, the same MEBN could be instantiated when dealing with varying number of entities.

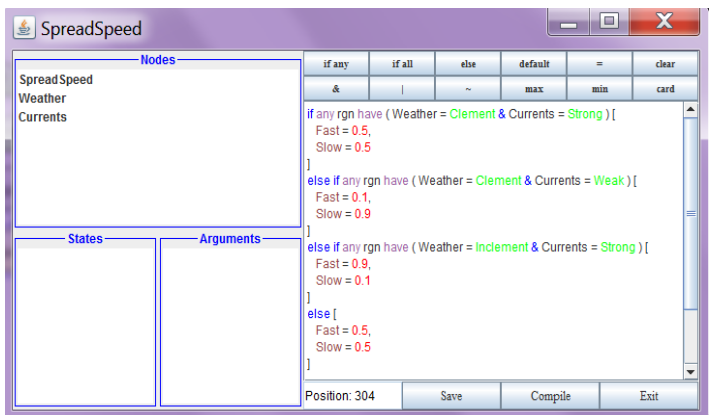

Fig. 4. The local probability distribution of SpreadSpeed.

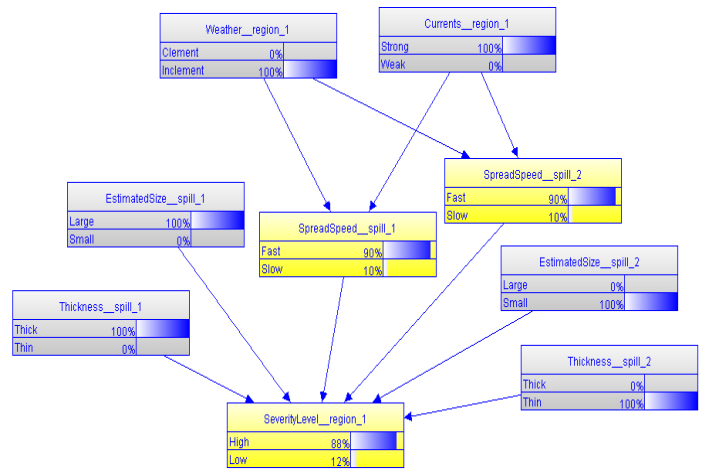

Fig. 5. Findings and query in the scenario.

Fig. 6. The SSBN for the scenario.

*D. Discussion*

Through this case study, the MEBN reasoning has shown its flexibility in adapting different situations over standard BN reasoning. It provides a compact way to represent repeated structures in a BN. Compared with BN, there is no fixed limit on the number of entity instances in the MEBN model, thus, it could allow dynamic instantiation of variables to accommodate the different situation. Another advantage of the MEBN reasoning is that MEBN models could be specified in an ontological language (e.g., OWL) using ontology constructs provided by the PR-OWL ontology. This feature is particularly attracting to the SWARMs project as ontology is the chosen solution to represent heterogeneous context in the cooperation of underwater vehicles. Thus, the MEBN reasoning could easily be applied to the SWARMs ontology and deduce more useful information. However, similar to the BN reasoning, the construction of MEBN models is highly dependent on the knowledge of domain experts. It implies huge demands for exhaustive and exclusive hypotheses. Also different domain expert might have a different interpretation for the same reasoning problem, thus, leading to the different modeling of MEBN for the same reasoning problem. In addition, to ensure a good approximation to the real probability distribution, a massive amount of historical data is needed.

## V. CONCLUSIONS

This paper has presented the current research conducted in the SWARMs project on facilitating context awareness in underwater robots. A conceptual proposal for context-aware framework has been presented. The proposed context-aware framework aims to provide a complete context management which offers different data treatments complying with the general lifecycle of context awareness. In the underwater robot field, context reasoning, as a significant phase to realize context awareness, is more difficult because context data obtained by robots are prone to be uncertain. To tackle the challenge in reasoning under uncertainties, the emerging probabilistic reasoning method, MEBN, has been studied and its applicability in the SWARMs project has been explored. The MEBN has been used to reason about the severity level of a spilled marine region in an oil spill monitoring scenario. The results have shown that the MEBN reasoning could be feasible to make inferences under uncertainties in such complex reasoning problems in underwater robots.

Future work would be focused on the following aspects:

- The presented context-aware framework is still at the prototypical level. To complete its implementation and testify it in real scenarios will be the next step forward.

- The MEBN reasoning will be tested with more use cases and the integration with the ontological and rule-based reasoning in the context reasoner is intended as future work.

- Context processing and reasoning are normally time and resource consuming. How to achieve a trade-off between the load on managing context data and the actual benefit obtained from high-level context needs to be figured out.

## *Acknowledgment*

The research leading to the presented results has been undertaken within the SWARMs European project (Smart and Networking Underwater Robots in Cooperation Meshes), under Grant Agreement n.662107-SWARMs-ECSEL-2014-1, partially supported by the ECSEL JU and the Spanish Ministry of Economy and Competitiveness (Ref: PCIN-2014-022-C02-02).

## *References*

[1] C. Perera, A. Zaslavsky, P. Christen, and D. Georgakopoulos, "Context-Aware Computing for The Internet of Things: A Survey," *IEEE Commun. Surv. Tutor.*, vol. 16, no. 1, pp. 414–454, 2014.

[2] X. Li, J.-F. Martínez, J. Rodríguez-Molina, and N. Martínez, "A Survey on Intermediation Architectures for Underwater Robotics," *Sensors*, vol. 16, no. 2, p. 190, Feb. 2016.

[3] K. B. Laskey, "MEBN: A language for first-order Bayesian knowledge bases," *Artif. Intell.*, vol. 172, no. 2–3, pp. 140–178, Feb. 2008.

[4] R. N. Carvalho, S. Matsumoto, K. B. Laskey, P. C. G. Costa, M. Ladeira, and L. L. Santos, "Probabilistic Ontology and Knowledge Fusion for Procurement Fraud Detection in Brazil," in *Uncertainty Reasoning for the Semantic Web II*, vol. 7123, F. Bobillo, P. C. G. Costa, C. d'Amato, N. Fanizzi, K. B. Laskey, K. J. Laskey, T. Lukasiewicz, M. Nickles, and M. Pool, Eds. Berlin, Heidelberg: Springer Berlin Heidelberg, 2013, pp. 19–40.

[5] R. N. Carvalho, R. Haberlin, P. C. G. Costa, K. B. Laskey, and K. Chang, "Modeling a Probabilistic Ontology for Maritime Domain Awareness," presented at the 2011 Proceedings of the 14th International Conference on Information Fusion (FUSION), Chicago, USA, 2011, pp. 1–8.

[6] G. Chantas *et al.*, "Multi-Entity Bayesian Networks for Knowledge-Driven Analysis of ICH Content," in *Computer Vision - ECCV 2014 Workshops*, vol. 8926, L. Agapito, M. M. Bronstein, and C. Rother, Eds. Cham: Springer International Publishing, 2015, pp. 355–369.

[7] The UnBBayes tool. Accessible online: http://unbbayes.sourceforge.net